# 4-Dimensional deformation part model for pose estimation using Kalman filter constraints

**Enrique Martinez Berti, Antonio Jose Sánchez Salmerón,
and Carlos Ricolfe Viala**

## Abstract

The goal of this research work is to improve the accuracy of human pose estimation using the deformation part model without increasing computational complexity. First, the proposed method seeks to improve pose estimation accuracy by adding the depth channel to deformation part model, which was formerly defined based only on RGB channels, to obtain a 4-dimensional deformation part model. In addition, computational complexity can be controlled by reducing the number of joints by taking into account in a reduced 4-dimensional deformation part model. Finally, complete solutions are obtained by solving the omitted joints by using inverse kinematic models. The main goal of this article is to analyze the effect on pose estimation accuracy when using a Kalman filter added to 4-dimensional deformation part model partial solutions. The experiments run with two data sets showing that this method improves pose estimation accuracy compared with state-of-the-art methods and that a Kalman filter helps to increase this accuracy.

## Keywords

DPM, Kalman filter, pose estimation, kinematic constraints, human activity recognition, computer vision, motion and tracking

## Introduction

Human pose estimation has been extensively studied for many years in computer vision. Many attempts have been made to improve human pose estimation with methods that work mainly with monocular RGB images.[1–5]

With the ubiquity and increased use of depth sensors, methods that use RGBD imagery are fundamental. One of the methods that used such imagery, and which is currently considered the state-of-the-art for human pose estimation, is Shotton et al.'s method,[6] which was commercially developed for the Kinect device. Shotton et al.'s method allows real-time joint detection for human pose estimation based solely on depth channel.

Despite the state-of-the-art performance of Shotton et al.'s method[6] and the commercial success of Kinect, the many drawbacks of Shotton et al.'s method[6] make it difficult to be adopted in any other type of 3-D computer vision system.

Some of the drawbacks of Shotton et al.'s algorithm[6] include copyright and licensing issues, which restrict the use and implementation of the algorithm for working on any other devices. Another drawback of the algorithm is the large number of training examples (hundreds of thousands) that are required to train its deep random forest algorithm and which could make training cumbersome.

Universitat Politecnica de Valencia, Instituto AI2, Valencia, Spain

**Corresponding author:**
Antonio Jose Sánchez Salmerón, Universitat Politecnica de Valencia, Instituto AI2, Camino de Vera s/n, Valencia, Spain.
Email: asanchez@isa.upv.es

Another drawback of Shotton et al.'s algorithm[6] is that its model is trained only on depth information and thus discards potentially important information that could be found in the RGB channels and could help approach human poses more accurately.

To alleviate these and other drawbacks in Shotton et al.,[6] we propose a novel approach that takes advantage of both RGB and depth information combined in a multichannel mixture of parts for pose estimation in single frame images coupled with a skeleton constrained linear quadratic estimator Kalman filter (SLQE KF) that uses the rigid information of a human skeleton to improve joint tracking in consecutive frames. Unlike Kinect, our approach makes our model easily trainable even for nonhuman poses. By adding depth information, we increase the time complexity of the proposed method. For this reason, we reduced the number of points modeled in the proposed method compared with the original deformation part model (DPM). Finally, to speed up the proposed method, we propose an inverse kinematics (IKs) method for the inference of the joints not considered initially, which cuts the training time.

The main contribution of our method extends to (i) an optimized multichannel mixture of parts model that allows the detection of parts in RGBD images; (ii) a linear quadratic estimator (LQE KF) that employs rigid information and connected joints of human pose; (iii) after adding depth information, time complexity was adversely affected. However, we could reduce the number of joints searched in our proposed method to overcome this inconvenience; and (iv) a model for unsolved joints through IK that allows the model to be trained with fewer joints and in less time.

Our results show significant improvements over the state-of-the-art in both the publicly available CAD60 data set and our own data set.

### Related work

Human pose estimation has been studied for many years, and some of the methods in the literature that attempt to solve this problem date back to the use of pictorial structures (PSs) introduced by Fischler and Elschlager.[7] More recent methods[3,8,9] improve the concept of PS with improved features or inference models.

Other methods that use more robust joint relationship include Yang and Ramanan's method[1] which uses a mixture of parts model, Sapp and Taskar's method[10] which, in turn, uses a multimodel decomposable model, and Wang et al.'s model[11] consider part-based models by introducing hierarchical poselets. Other methods that have attempted to reconstruct 3-D pose estimation from RGB monocular images include the methods of Bourdev and Malik,[12] Ionescu et al.,[13] and Gkioxari et al.[14]

Object detection has been done using RGBD with Markov Random Fields (MRFs) and features from both RGB and depth.[15]

Recently, 3-D cameras such as Kinect have added a new dimension to computer vision problems. Such cameras allow us to capture not only RGB information as done with monocular cameras but also depth information whose intensities depict an inversely proportional relationship of the distance of the objects to the camera.

Some methods that use depth images to reconstruct pose estimations include the methods of Grest et al.,[16] Plagemann et al.,[17] Shotton et al.,[6] Helten et al.,[18] Baak et al.,[19] and Spinello and Arras.[20] Among such methods, Shotton et al.'s method,[6] which was developed for the Kinect algorithm, has become the state-of-the-art for performing human pose estimation that predicts 3-D positions of body joints from a single depth image.

### Proposed method

In this section, we first explain the preprocessing step for the depth channels in which the background was removed to improve the accuracy of our algorithm (see Figure 1). The "Multichannel mixture of parts" section explains the formulation of our 4-D mixture of parts model. The "Joint detection in consecutive frames" section explains our structured LQE for correcting joints in consecutive frames. Finally, the "Model simplification" section describes the strategy to reduce the computational complexity of our proposed method.

### Data preprocessing

As a processing step of RGB channels, we isolate significant foreground areas in these channels from background noise. This is done by removing regions in the depth images that are most stable to different thresholds that belong to the background. Such a foreground and background template is then transferred to the RGB images to thus remove noise or conflicting object patterns that would confuse foreground and background features in our method and would hinder detection accuracies.

The intuition behind this approach is that objects or people in the foreground seen through the depth sensor share areas with similar pixel intensities. The reason for this is that the infrared (IR) rays being reflected from the objects in the foreground are reflected more or less at the same time and with the same intensity. Other objects or areas that are much farther away from the IR camera unevenly reflect such rays, and these areas appear more noisy and with varying intensities. Figure 2 shows the different intensities reflected from the IR sensor that represents the depth coordinates of the objects.

Due to this property of the pixel intensities in the depth images, our background removal method, which is used for depth and later applied to the RGB images, uses a maximally stable extremal region (MSER)-based approach.[21] These regions are the most stable ones within a range of all possible threshold values being applied to them. A stability score $\delta$ of each region in the depth channels is calculated so that $\delta = \frac{|\Delta R - R|}{|R|}$, where $|R|$ represents the area of the region in question and $\Delta$ represents the intensity variation for the different thresholds. Hence, we remove those MSERs in which areas are above a $T$ threshold. We train the parameters for MSER based on a subset of the training set. We can see in Figure 7 the results from our background subtraction method. Note that most of the noisy pixels in the background have been removed.

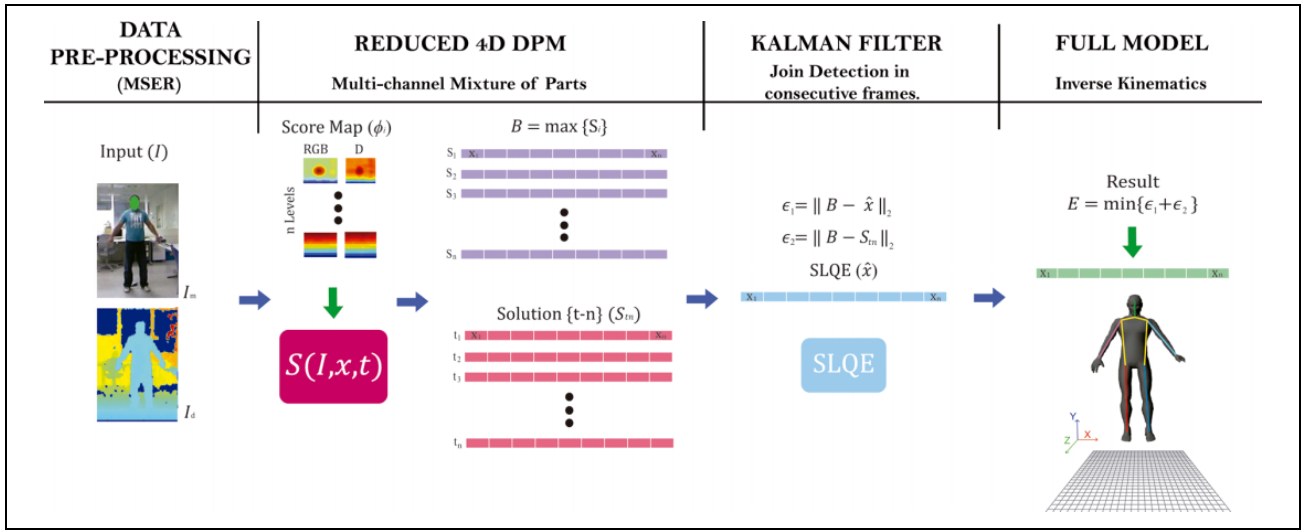

**Figure 1.** Outline of our method.

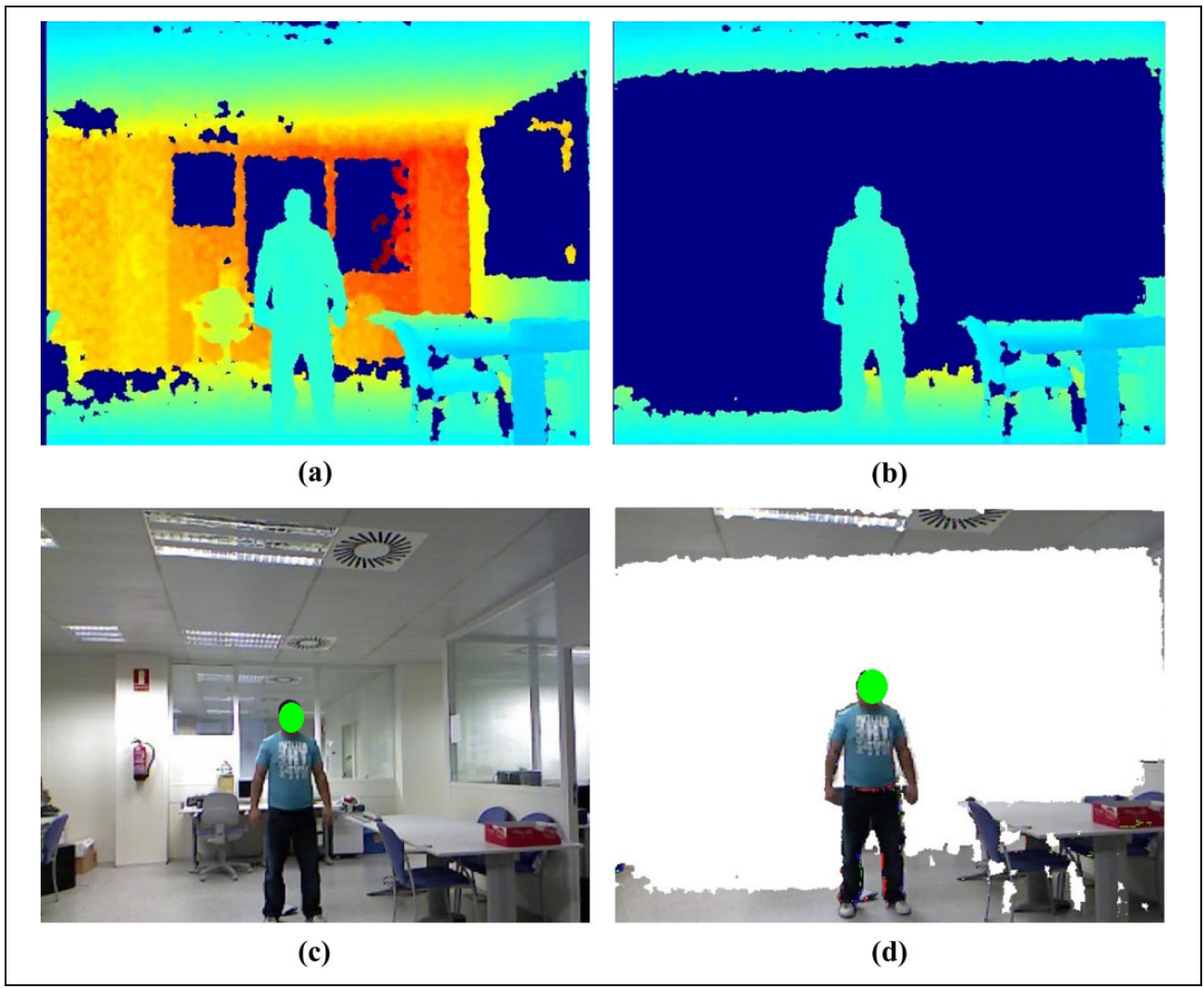

**Figure 2.** (a) Original depth; (b) depth after applying MSER; (c) original RGB; (d) combining images (c) and (d). MSER: maximally stable extremal region.

## Multichannel mixture of parts

Until recently, Yang and Ramanan's method[1] has been a state-of-the-art method for pose estimation in monocular images. Yet as we can see in Figure 6 of our "Results" section, Yang and Ramanan's method performs poorly on images that vary from those in its training set, and their method only improves by a small margin even after retraining.

Although there have been other algorithms[2,3,5] that have improved Yang and Ramanan's model, all these methods, including Yang and Ramanan's, use a mixture of parts for only the RGB dimension of channels. Conversely, our method uses a multichannel mixture of parts model that allows us to extend the number of mixtures of parts to the depth dimension of *RGBD images*.

The depth channel increases time complexity, but this disadvantage has been solved by cutting the number of joints modeled in our 4-dimensional DPM (4D-DPM) method. Hence, our method differs significantly from other previous methods in many important ways that we explain in this section.

In our method, we formulate a score function ($S$) for the parts or joints that belong to pose through an appearance and deformation functions as follows[1]

$$S(I, x, t) = \sum_{i \in V} \phi_i(I, x_i, t_i) + \sum_{ij \in E} \psi_{i,j}(I, x_i, t_i, x_i') \quad (1)$$

where $I$ corresponds to the RGBD image, $x$ is the location of joint $i$, which corresponds to the type of joint being detected, $j$ is the potential joint being connected to $i$ and $t = 1, \ldots, T$ is the mixture component of joint $i$ that expands to parts that have undergone different transformations, such as rotation, translation, orientation, and others, and where $x_i' = (x_j, t_j)$. The terms $\phi$ and $\psi$ in equation (1) correspond to appearance model and deformation model, respectively. The appearance model calculates a score for the features of type assignment $t_i$, whereas the deformation model provides a score for the deformation distance of type assignments $t_i$ and $t_j$. These models are constrained with the tree structure represented by $G(V, E)$, where a vertex $i \in V$ represents a part and the edge $(i, j) \in E$ denotes the co-occurrence of parts $i$ and $j$ for optimization purposes because the computation time of all the possible assignments is exponential.

In order to obtain features and deformations in all RGBD channels, we formulate $\phi$ and $\psi$ as a multichannel mixture of parts in the following way

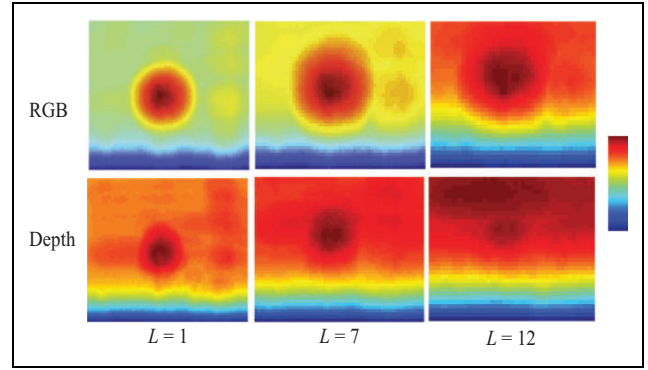

**Figure 3.** Score maps of component at different levels. The figure shows that mixture of parts in RGBD is complementary.

$$\phi_i(I, x_i, t_i) = \begin{bmatrix} \omega_{i\,m}^{t_i} \cdot \phi(I_m, x_i) + b_{i\,m}^{t_i} \\ \omega_{i\,d}^{t_i} \cdot \phi(I_d, x_i) + b_{i\,d}^{t_i} \end{bmatrix}$$

$$\psi_{ij}(I, x_i, t_i, x_j, t_j) = \begin{bmatrix} \omega_{ij\,m}^{t_i,t_j} \cdot \psi(x_i - x_j)_m + b_{ij\,m}^{t_i,t_j} \\ \omega_{ij\,d}^{t_i,t_j} \cdot \psi(x_i - x_j)_d + b_{ij\,d}^{t_i,t_j} \end{bmatrix} \quad (2)$$

where $\phi(I, x_i)$ is the appearance function represented by Histogram of Gradients (HOG)[22] that extracts features from monocular ($I_m$) or depth ($I_d$) images at pixel location $x_i$. $m$ represents a monocular part and $d$ denotes a depth part. $\omega$ are the previously trained filters. $b_i^{t_i}$ is a parameter that corresponds to the assignment of part $i$ in either channel and $b_{ij}^{t_i,t_j}$ is another parameter that describes the co-occurrence assignments of parts $i$ and $j$. Note that, unlike Yang and Ramanan,[1] the number of mixture parts in our equation (2) is twice as many because a depth channel is added. This extra number of mixture components is a complement to mixtures from RGB dimensions and allows to improve the detection scores for all RGBD channels. This property is also seen in Figure 3, which shows the different scores collected from different channels.

The deformation function is given by $\psi(x_i - x_j)_c = \begin{bmatrix} dx & dx^2 & dy & dy^2 \end{bmatrix}$, where $dx = x_i - x_j$ and $dy = y_i - y_j$, which correspond to the location of part $i$ compared to $j$ in image $I_c$ for the respective type of image $c$.

As the structure of $G(V, E)$ is a tree, we use dynamic programming to calculate the $S$ for each node in the tree with an extra second term compared to Yang and Ramanan[1] to calculate the scores and message passing in a way to accommodate for depth channels. Let kids($i$) be the set of children of part $i$ in $G$. We compute the message part $i$ that passes to its parent $j$ in this way

$$\text{score}_i(t_i, x_i) = b_i^{t_i} + \begin{bmatrix} \omega_{i\,m}^i \cdot \phi(I_m, p_i) \\ \omega_{i\,d}^i \cdot \phi(I_d, p_i) \end{bmatrix} + \sum_{k \in \text{kids}(i)} m_k(t_i, x_i) \quad (3)$$

$$m_i(t_j, x_j) = \max_{t_i} b_{ij}^{t_i, t_j} \max_{x_i} \text{score}(t_i, x_i)$$
$$+ \begin{bmatrix} w_{ij}^{t_i, t_j} m \cdot \psi(x_i - x_j)_m \\ w_{ij}^{t_i, t_j} d \cdot \psi(x_i - x_j)_d \end{bmatrix} \quad (4)$$

Equation (3) computes the local score of part $i$, at all the pixel locations $p_i$ and for all possible types $t_i$, by collecting messages from the children of part $i$. Equation (4) computes every location and type of its child part $i$. Once messages are passed to the root ($i = 1$), $\text{score}_1(c_1, x_1)$ represents the best scoring configuration for each root type and position.

In contrast to Yang and Ramanan,[1] we parametrize equation (1) as $S(I, x, t) = \alpha \cdot \Phi(I, x, t)$ and $\alpha = (w, b)$ to solve the following structural support vector machine primal with the following conditions for processing positive and negative samples, which allows us to solve the most violated constraint as independent steps $i$ and to thus improve training times compared to Yang and Ramanan[1]

$$\arg\min_{w, \xi \geq 0} \frac{1}{2} \alpha \cdot \alpha + C \sum_n \xi_n$$
$$\text{s.t. } \forall n \in \text{pos } \alpha \cdot \Phi(I_{ni}, x_{ni}, t_{ni}) \geq 1 - \xi_{ni} \quad (5)$$
$$\forall n \in \text{neg}, \forall x_n, t_n \, \alpha \cdot \Phi(I_n, x_n, t_n) \leq -1 + \xi_n$$

### Joint detection in consecutive frames

To date, we have dealt only with pose estimation for each single frame independently. However, most of the joint movement performed in normal circumstances displays uniform and constant changes of displacement and velocity. Hence, we can use the properties of the velocity and acceleration of joints to make predictions based on the past where joints would most likely be. This motion-based prediction could help us to validate our frame-based prediction.

One way of predicting joint location based on previous detections is by using an LQE KF.[23] Using a simple LQE works well when the joints being tracked are independent of each other and their movement does not correlate. However, in our case, our joints are connected to each other through limbs, which are rigid connections and allow the movement of one joint related to the other one to be connected; for example, the foot joint movement would be relative to a parent joint such as a knee or a hip.

In order to utilize this joint relationship, we introduce a novel SLQE, which uses joint relationship constraints from a human skeleton model to predict the location of joints at the same time. In this section, we explain this step of our approach.

We first define a state joint obtained by equation (6) with its respective vector components for position $(x_i, y_i)$, velocity $(vx_i, vy_i)$, and acceleration $(ax_i, ay_i)$ as follows

$$x_i' = \begin{bmatrix} x_i & y_i & vx_i & vy_i & ax_i & ay_i \end{bmatrix}^T \quad (6)$$

We also define the measurement matrix for a joint as $H_1$ that considers only location components $x_i$ and $y_i$ of the joint

$$H_1 = \begin{bmatrix} 1 & 0 & 0_{1 \times 4} \\ 0 & 1 & 0_{1 \times 4} \\ 0_{4 \times 1} & 0_{4 \times 1} & 0_{4 \times 4} \end{bmatrix}_{6 \times 6} \quad (7)$$

Thus, the measurement matrix for all the joints is represented as

$$H = \begin{bmatrix} H_1 & 0_{6 \times 6} & 0_{6 \times 6} & 0_{6 \times 6} \\ 0_{6 \times 6} & H_1 & 0_{6 \times 6} & 0_{6 \times 6} \\ \vdots & \vdots & \ddots & \vdots \\ 0_{6 \times 6} & 0_{6 \times 6} & 0_{6 \times 6} & H_1 \end{bmatrix}_{48 \times 48} \quad (8)$$

Given a state model $A$, which models the relationship of each joint to all the other joints being considered, we define a pair of joints that are connected to each other as $A_1$ and $A_2$ to be

$$A_1 = \begin{bmatrix} 1 & 0 & 1 & 0 & 0 & 0 \\ 0 & 1 & 0 & 1 & 0 & 0 \\ 0 & 0 & 1 & 0 & 1 & 0 \\ 0 & 0 & 0 & 1 & 0 & 1 \\ 0 & 0 & 0 & 0 & 1 & 0 \\ 0 & 0 & 0 & 0 & 0 & 1 \end{bmatrix}_{6 \times 6} \quad (9)$$

where the main diagonal represents the same elements as equation (6) and the upper diagonal denotes the relationships between these elements (e.g. $vx_i$ to depend on $x_i$). We take 1 to describe these relationships

$$A_2 = \begin{bmatrix} 0 & 0 & -1 & 0 & 0 & 0 \\ 0 & 0 & 0 & -1 & 0 & 0 \\ 0 & 0 & 0 & 0 & -1 & 0 \\ 0 & 0 & 0 & 0 & 0 & -1 \\ 0 & 0 & 0 & 0 & 0 & 0 \\ 0 & 0 & 0 & 0 & 0 & 0 \end{bmatrix}_{6 \times 6} \quad (10)$$

where the upper diagonal represents how the relationships in the consecutive frames change. By changing this value, we can change the velocity of the predicted joints, and to what extent a point, compared to a previous one, can be predicted. After some experiments, we took $-1$ to represent velocity in the system changes

$A_1$ is fixed and $A_2$ can be adjusted to fast track the movement dynamics. Thus, the final transition state matrix $A$ for all the joints is defined as

$$A = \begin{bmatrix} A_1 & A_2 & 0 & 0 & 0 & 0 & 0 & 0 \\ 0 & A_1 & 0 & 0 & 0 & A_2 & 0 & 0 \\ 0 & 0 & A_1 & 0 & 0 & 0 & A_2 & 0 \\ 0 & 0 & A_2 & A_1 & 0 & 0 & 0 & 0 \\ 0 & 0 & 0 & 0 & A_1 & A_2 & 0 & 0 \\ 0 & 0 & 0 & 0 & 0 & A_1 & 0 & 0 \\ 0 & 0 & 0 & 0 & 0 & 0 & A_1 & 0 \\ 0 & 0 & 0 & 0 & 0 & 0 & A_2 & A_1 \end{bmatrix}_{48 \times 48} \quad (11)$$

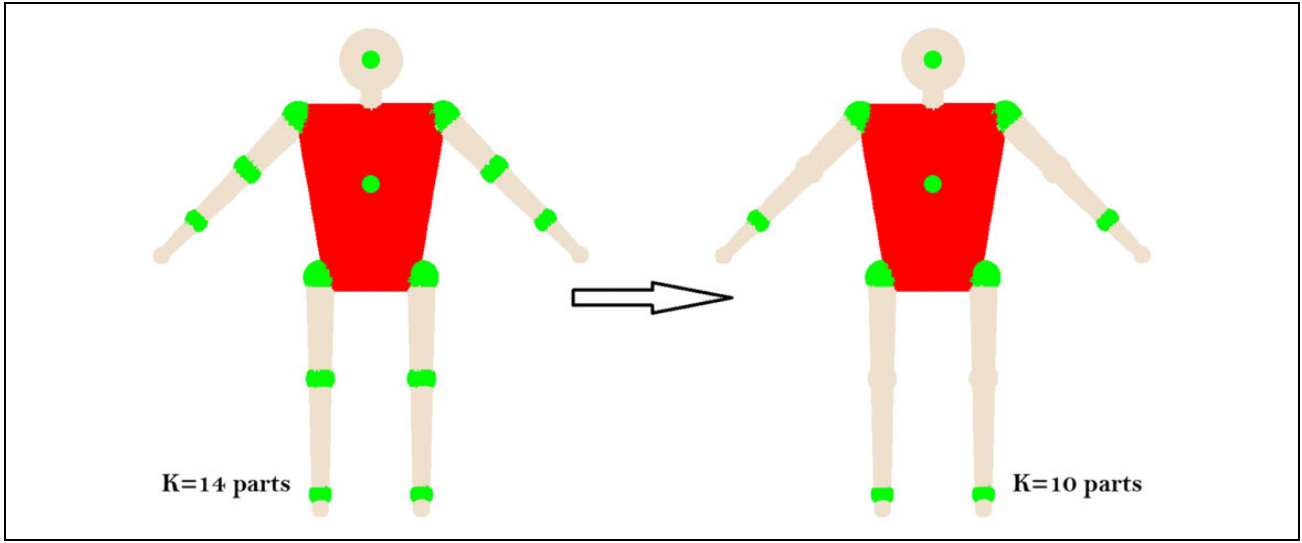

**Figure 4.** Left: full model with 14 parts (green points). Right: reduced model with 10 parts.

Note that the joints whose movement depends on another joint are paired up through the relationship $A_1A_2$. The movement of joints that are connected to each other is dependent on each other, thus their velocity and acceleration components are subtracted from each other. Matrix $A$ represents our observed model that is to be predicted. Choosing the correct matrix $A$ is important to correctly predict joints.

The prediction of a posteriori joint $\mathbf{x} = [x'_1, \ldots, x'_n]$ at time $t$ now depends on the structure embedded in $A$ and can be calculated with

$$\mathbf{x}_t = \mathbf{A}\mathbf{x}_{t-1} \tag{12}$$

We also calculate a posteriori error covariance $P_t$ so that

$$P_t = AP_{t-1}A^T + Q \tag{13}$$

where $Q$ is the measurement noise, which is an identity matrix in our case.

We also compute residual covariance $S$ based on noise covariance prediction $R$ to calculate gain $K$ in this way

$$\begin{aligned} S &= HP_tH^T + R \\ K &= P_tH^TS^{-1} \end{aligned} \tag{14}$$

Once the outcome of measurement $\mathbf{x}$ is obtained, these estimates are updated using gain $K$, but with more weight being given to the estimates with greater certainty.

The final estimation of the coordinate joints by our SQLE is given by

$$\hat{x} = H \cdot \mathbf{x}_{t-1} \tag{15}$$

Although SLQE can accurately predict the direction and speed of movement for continuous movements, in these cases, joint movement changes direction suddenly, so prediction can fail.

To avoid this issue, we compare our prediction from SLQE and the last successful prediction from the last frame $B = \max_i S_{it}$, where $S_i$ is the score function from 1 at frame $t$.

Thus, we can avoid making mistakes by SQLE or the score function by choosing the solution $\hat{x}$ or $S_{t-1}$ with the least error $\min(\varepsilon_1, \varepsilon_2)$

$$\begin{aligned} \epsilon_1 &= \parallel B - \hat{x} \parallel_2 \\ \epsilon_2 &= \parallel B - S_{t-1} \parallel_2 \end{aligned} \tag{16}$$

Given the algorithm's recursive nature, this process can run in real time using only the present input measurements and the previously calculated state and its uncertainty matrix. No additional past information is required.

### 3-D pose estimation

Once the coordinates of joints have been calculated in planes $X$ and $Y$, finding their coordinates in the $Z$ plane is as simple as converting the pixel values into the depth images and back into $Z$ coordinates.

### Model simplification

The additional depth images included in our formulation add a computational cost to our training and testing phases.

In this section, we explain a simplification technique that uses inverse kinematic equations in order to infer shoulder and knee joints. The original DPM calculates the full body parts with 14 joints. By using IKs, we can lower that number of points to 10. The joints modeled in our proposed 4D-DPM method were reduced, as were the variables to be predicted with KF.

Figure 4 shows the full model with 14 parts on the left and the reduced model with 10 parts on the right, where the joints from the elbow and knee have been deleted.

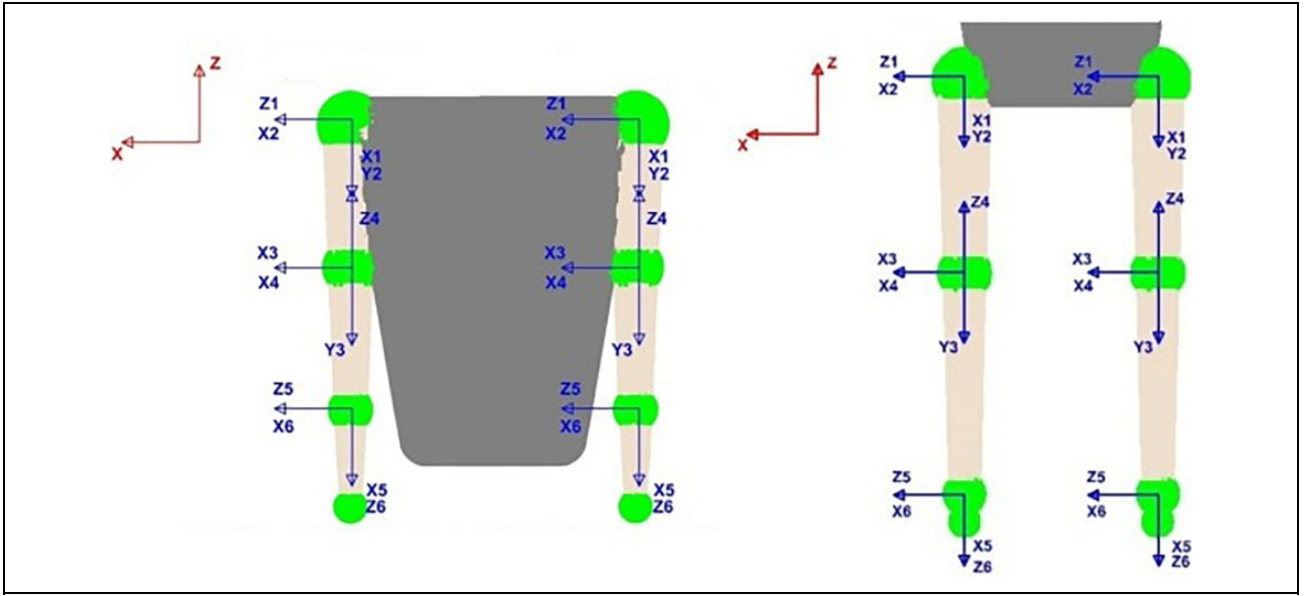

**Figure 5.** State variables. Left: coordinate systems of the arms. Right: coordinate systems of the legs.

*Human body model.* In order to track the human skeleton, we model it as a group of kinematic chains, where each part and joint in the human body corresponds to a link and joint in a kinematic chain. Given the joint positions predicted by the KF, IKs are used to obtain full joints using Denavit–Hartemberg (D-H) model.[24,25]

*State variables.* The human body model is divided into four kinematic chains (KCs), namely in essence, one KC for each arm and one KC for each leg.

Figure 5 shows the coordinate system for each part used to represent legs and arms. The reduced model uses only shoulder and hand points to represent arms, and hip and feet to represent legs. However, by using the IKs with the coordinate systems described in Figure 5, we can obtain elbow and knee points and obtain the full model with 14 points. All these coordinate systems are represented in relation to the same base coordinate system. Since the proposed 4D-DPM method returns the relationships of the locations between all the parts, each KC can be considered independent of the others.

*D-H model.* We use D-H to model each KC. Hence, we use six joints for each KC for shoulders, hips, hands, and feet (see Figure 5).

First, we establish the base coordinate system $(X_0, Y_0, Z_0)$ at the supporting base with the $Z_0$-axis lying along the axis of motion of joint 1. Then, we establish a joint axis and align the $Z_i$ with the axis of motion of joint $i + 1$.

We also locate the origin of the $i$-th coordinate at the intersection of the $Z_i$ and $Z_{i-1}$ or at the intersection of a common normal between the $Z_i$ and the $Z_{i-1}$. Then, we establish $X_i = \pm (Z_{i-1} \times Z_i)/||Z_{i-1} \times Z_i||$ or along the common normal between the $Z_i$- and $Z_{i-1}$-axes when they are parallel. We also assign $Y_i$ to complete the right-handed coordinate system. Finally, we find the link and joint

**Table 1.** D-H table.

| | $\theta$ (deg) | $d$ (mm) | $\alpha$ (deg) | $a$ (mm) |
|---|---|---|---|---|
| $q_1$ | $\theta_1$ | 0 | $\alpha_1$ | 0 |
| $q_2$ | $\theta_2$ | 0 | $\alpha_2$ | 0 |
| $q_3$ | $\theta_3$ | $d_3$ | $\alpha_3$ | $a_3$ |
| $q_4$ | $\theta_4$ | 0 | $\alpha_4$ | 0 |
| $q_5$ | $\theta_5$ | $d_5$ | $\alpha_5$ | $a_5$ |
| $q_6$ | $\theta_6$ | 0 | $\alpha_6$ | 0 |

$\theta_i$: rotation along axis $Z_{i-1}$ to put axis $X_{i-1}$ on axis $X_i$; $\alpha_i$: rotation along axis $X_i$ to put axis $Z_{i-1}$ on axis $Z_i$; $d_i$: translation between coordinate system $O_{i-1}$ and $O_i$ along axis $Z_{i-1}$; $a_i$: translation between the coordinate system $O_{i-1}$ and $O_i$ along axis $X_i$.

parameters: $\theta_i$ (angle of the joint compared to the new axis), $d_i$ (offset of the joint along the previous axis to the common normal), $a_i$ (length of the common normal), and $\alpha_i$ (angle of the common normal compared to the new axis).

For each KC, we have six variable joints $q_i$. Each $q_i$ is placed on the $z_i$-axis in Figure 5. Now, we can define the table of the D-H parameters. A generic D-H parameter table for the proposed KC is shown in Table 1. Given the six variable joints $(q_1, q_2, q_3, q_4, q_5, q_6)$, we obtain the coordinates of end effector $(x, y, z)$ compared to the base of KC. For IKs, given the coordinates of the end effector and the orientation in Euler parameters $(x, y, z, \phi, \theta, \psi)$, we obtain the six variable joints $(q_1, q_2, q_3, q_4, q_5, q_6)$.

Given the homogeneous transformation matrix that establishes the relationship of a joint with an adjacent one

$$^{i-1}A_i(q_i) = \begin{bmatrix} c_\theta & -c_\alpha \cdot s_\theta & s_\alpha \cdot s_\theta & a_i \cdot c_\theta \\ s_\theta & c_\alpha \cdot c_\theta & -s_\alpha \cdot c_\theta & a_i \cdot s_\theta \\ 0 & s_\alpha & c_\alpha & d_i \\ 0 & 0 & 0 & 1 \end{bmatrix} \quad (17)$$

where $s_\alpha = \sin(\alpha_i)$, $c_\alpha = \cos(\alpha_i)$, $s_\theta = \sin(\theta_i)$, $c_\theta = \cos(\theta_i)$, and $\alpha, \theta, d, a$ are the DH parameters.[26,27] The location of the end effector in relation to the reference can be obtained by the following relationship

$$^0T_6(q_1, q_2, q_3, q_4, q_5, q_6) = {}^0A_1 \cdot {}^1A_2 \cdot {}^2A_3 \cdot {}^3A_4 \cdot {}^4A_5 \cdot {}^5A_6$$

where $A_i = {}^{i-1}A_i(q_i)$. It is paramount to use geometric models for the first three joints. Thus, we obtain the coordinates for final effector $(x, y, z)$ and, after applying geometric models, we can obtain the first three joints

$$q_1 = \arctan\left(\frac{y}{x}\right) \tag{18}$$

$$q_3 = \arctan\left(\frac{\pm\sqrt{1 - \cos^2\left(\frac{x^2+y^2+z^2-a_2-a_3}{2 \cdot a_2 \cdot a_3}\right)}}{\cos\left(\frac{x^2+y^2+z^2-a_2-a_3}{2 \cdot a_2 \cdot a_3}\right)}\right) \tag{19}$$

$$q_2 = \arctan\left(\frac{z}{\pm\sqrt{x^2+y^2}}\right) - \varphi \tag{20}$$

where

$$\varphi = -\arctan\left(\frac{a_3 \cdot \sin\left(\frac{x^2+y^2+z^2-a_2-a_3}{2 \cdot a_2 \cdot a_3}\right)}{a_2 + a_3 \cdot \cos\left(\frac{x^2+y^2+z^2-a_2-a_3}{2 \cdot a_2 \cdot a_3}\right)}\right)$$

Now, we can use IKs to calculate the last three joints. We define $^0R_6 = {}^0R_3 \cdot {}^3R_6$ for the submatrix rotation of $^0T_6$. We know the value of $^0R_6$ because it is the orientation of the final effector and $^0R_3$ because it is defined by $^0R_3 = {}^0R_1 \cdot {}^1R_2 \cdot {}^2R_3$ using $(q_1, q_2, q_3)$. Then we calculate

$$^3R_6 = [r_{ij}] = ({}^0R_3)^{-1} \, {}^0R_6 \tag{21}$$

By applying $^3R_6 = {}^3R_4 \cdot {}^4R_5 \cdot {}^5R_6$ and using $(q_4, q_5, q_6)$, we obtain the last three joints using equation (21)

$$q_4 = \arctan\left(\frac{r_{23}}{r_{13}}\right) \tag{22}$$

$$q_5 = \arccos(-r_{33}) \tag{23}$$

$$q_6 = \frac{\pi}{2} - \arctan\left(\frac{r_{32}}{r_{31}}\right) \tag{24}$$

We use IKs because we can obtain the base of our KC (shoulders or hips), and where the final effector and orientation (hands and feet) are, thus we obtain these parameters $(x, y, z, \phi, \theta, \psi)$. Using IKs, we obtain the six variable joints $(q_1, q_2, q_3, q_4, q_5, q_6)$ and use them to know where the elbow or knee is located.

Figure 6 shows at the top the solutions from the proposed method using 10 parts. These parts correspond to the 10 parts shown in Figure 4 on the right. The bottom images show the full model solutions after applying IKs.

## Results

### 3-D camera calibration

Our method works with any RGBD sensor after correct calibration. In our experiments, we use a Kinect device and calibrate the intrinsic and extrinsic parameters of the monocular and IR sensors. The calibration system is done similarly to Berti et al.[27] or Viala et al.[28,29]

### Data sets

To train and test our method, we use a combination of videos from our own data set and a subset of the publicly available CAD60 data set.[30]

### CAD60 data set

The original CAD60 data set.[30] contains 60 RGB-D videos, 4 subjects (2 male, 2 female), 4 different environments (office, bedroom, bathroom, and living room), and 12 different activities. This data set was originally created for the activity recognition task.[31,32,33] The size of the images is 320×240 pixels.

### Our data set

It consists of seven videos with only one person on the scene moving his arms and legs. We had almost 1000 frames of people to obtain specific movements, for example crossing arms over one's body, to complement the CAD60 data set. Images were taken indoors in different scenarios. The subject inside the images is male who wears different clothes. The size of the images is 320×240 pixels.

The ground truth of the joints in this data set was obtained by recording predictions from Kinect. Thus, in order to make a fair comparison of the predictions from the methods being tested, we provide the videos to our human annotators to manually record the ground truth of the joint positions in the CAD60 data set. Thus, our annotators recorded over 15,000 frames of videos that correspond to 16 videos from the CAD60 data set with different activities and environments. For training and testing purposes, we use two different splits of such annotations. We chose to manually annotate the CAD60 data set because, to our knowledge, there is no RGBD data set with ground truth of human pose joints. We will also publicly release our annotated videos for the benefit of the research community.

### Metrics

The metrics we use in our different experiments are probability of a correct keypoint (PCK), Average Precision Keypoint (APK), and error distance.

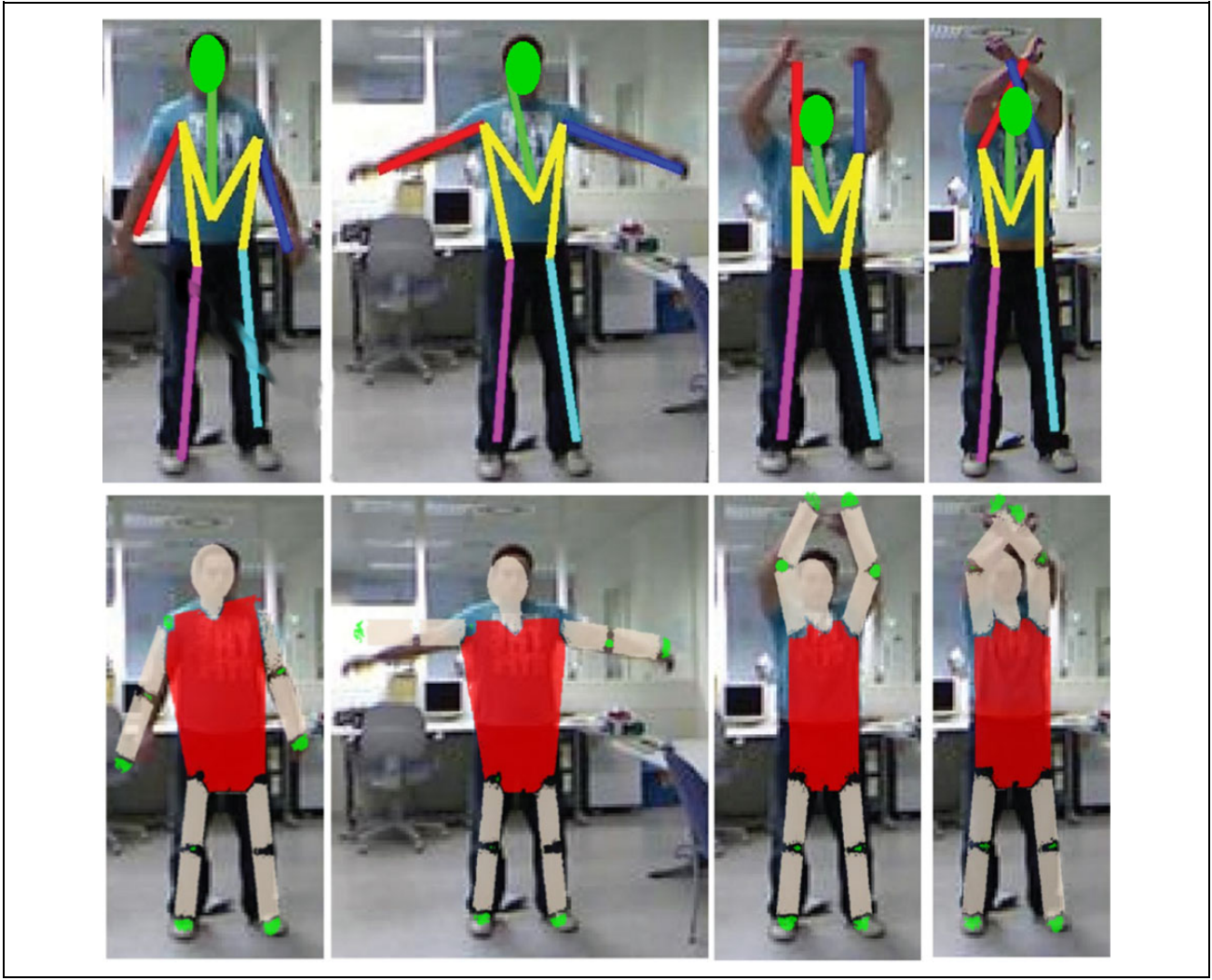

**Figure 6.** Results of our method. First row shows joints of the reduced model on a sequence which does not belong to CAD60 data set. Second row shows the full model inferred where elbows and knees are estimated by IKmodel. IKs: inverse kinematics.

**Table 2.** Experimental comparisons with the state-of-the-art methods and different components of our methods on CAD60 data set.[a]

| Model | Metric | Head | Shoulder | Wrist | Hip | Ankle | Average |
|-------|--------|------|----------|-------|-----|-------|---------|
| Yang (Yang and Ramanan[1]) | APK | 47.30 | 66.70 | 22.40 | 45.50 | 47.10 | 46.50 |
| | PCK | 62.50 | 70.40 | 39.00 | 60.50 | 57.9 | 58.06 |
| | Error | 15.53 | 12.23 | 22.34 | 16.29 | 18.50 | 16.97 |
| Kinect (Shotton et al.[35]) | APK | 68.30 | 90.70 | 76.40 | 9.50 | 77.10 | 64.40 |
| | PCK | 79.50 | 94.40 | 85.00 | 23.50 | 85.9 | 73.66 |
| | Error | 13.17 | 6.85 | 9.64 | 18.42 | 11.28 | 15.87 |
| P. Method | APK | 72.30 | 91.10 | 81.20 | 83.70 | 82.00 | 82.06 |
| | PCK | 83.60 | 95.00 | 88.70 | 87.30 | 89.20 | 88.76 |
| | Error | 9.95 | 6.81 | 8.73 | 8.58 | 8.40 | 8.49 |

PCK: probability of a correct keypoint.
[a]APK and PCK metrics are expressed in percent. Error is expressed in pixels. Italics represent higher values.

## PCK

The PCK was introduced by Yang and Ramanan.[1] Given the bounding box, a pose estimation algorithm must report back the keypoint locations for body joints. The overlap between the keypoint bounding boxes was measured, which can suffer from quantization artifacts for small bounding boxes. A keypoint is considered correct if it lies within $\alpha \cdot \max(h, w)$ of the ground truth bounding box, where $h$ corresponds to the

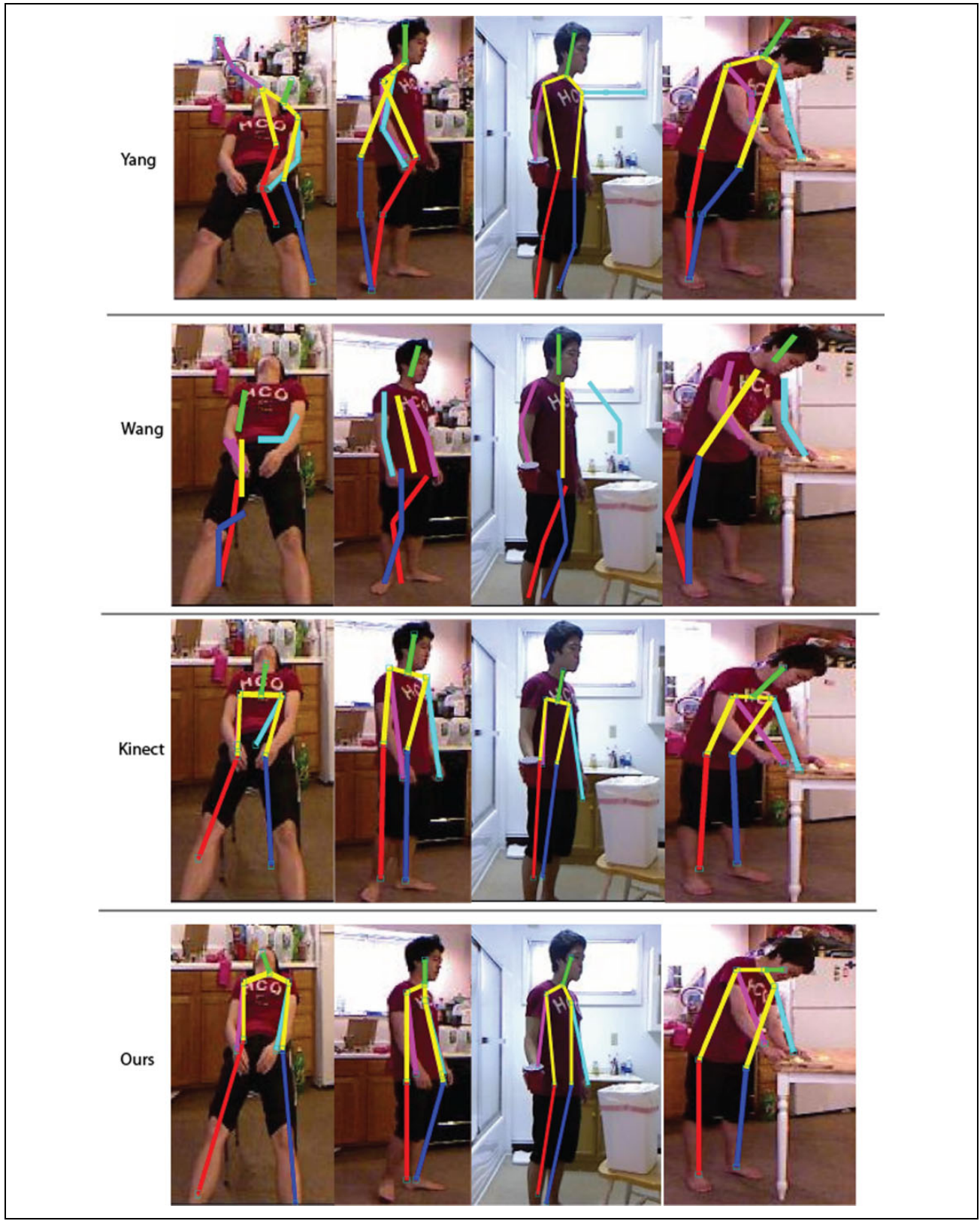

**Figure 7.** Qualitative comparison of four different methods for pose estimation on four sequences which belong to CAD60 data set. Fourth row shows joints of the reduced model.

**Table 3.** Experimental comparisons with the state-of-the-art methods on our proposed data set.[a]

| Model | Metric | Head | Shoulder | Wrist | Hip | Ankle | Average |
|---|---|---|---|---|---|---|---|
| Yang (Yang and Ramanan[1]) | APK | 92.20 | 92.30 | 82.70 | 86.60 | 83.50 | 87.26 |
| | PCK | 91.50 | 89.00 | 85.80 | 89.90 | 83.80 | 88.00 |
| | ERROR | 8.17 | 8.81 | 10.87 | 9.37 | 11.59 | 9.76 |
| P. Method (without KF) | APK | 94.20 | 95.10 | 88.30 | 89.70 | 90.30 | 91.52 |
| | PCK | 93.80 | 92.50 | 88.90 | 90.30 | 91.00 | 91.30 |
| | ERROR | 6.48 | 6.02 | 8.73 | 8.01 | 7.66 | 7.38 |
| P. Method* (with KF) | APK | 97.50 | 98.30 | 92.20 | 94.70 | 94.00 | 95.34 |
| | PCK | 96.40 | 95.20 | 93.70 | 96.50 | 94.20 | 95.20 |
| | ERROR | *5.82* | *5.71* | *7.43* | *6.37* | *6.61* | *6.38* |

PCK: probability of a correct keypoint.
[a]APK and PCK metrics are expressed in percent. Error is expressed in pixels.
*Signifies difference between two equals methods trained differently. Italics represent higher values.

height and $w$ to the width of the corresponding bounding box and $\alpha$ is a parameter that controls the relative threshold to consider the correctness of the keypoint.

## APK

In a real system, however, one has no access to annotated bounding boxes at the test time, and one must also address the detection problem. One can cleanly combine the two problems by thinking of body parts (or rather joints) as objects to be detected and evaluate object detection accuracy with a precision–recall curve. The average precision keypoint is another metrics introduced by Yang and Ramanan,[1] where, unlike PCK, it penalizes false-positives. Correct keypoints are also determined through the $\alpha \cdot \max(h, w)$ relationship.

## Error distance

This metrics calculates the distance between the results and the correct labeled point. To do this, we calculate the distance error between the predicted result and the ground truth location. For each joint, we obtain an error score that is the mean value calculated from all the frames.

## Quantitative results

Table 2 shows the results of comparing our proposed method (P. Method) with other methods, such as Shotton et al.'s method,[6] which is used with the Kinect device. Some of the issues we encountered with the Kinect algorithm is that the detections which vary from frame to frame are not consistent. Moreover, Kinect usually mis-predicts hip joints compared to our ground truth, which was generated by our human annotators. We can also see in Figure 7 that Kinect has issues with correctly positioning head, ankle, and wrist joints.

Although a fairer comparison with Shotton et al.[6] would be to use the exact training set for both algorithms, such a comparison of the training step is difficult to make because there is no open source of the Kinect algorithm available to produce this type of experiments.

Unlike Shotton et al.'s method,[6] in our experiments we observe that our algorithm can produce competitive results, even with only a few hundred frames in the CAD60 training set.

We also compare our results with Yang and Ramanan's[1] original method trained on the image parse data set[34] in Table 2 and also retrain it (Yang*) with the same images that we used to train our proposed method (P. Method*; Table 3). Note that although we retrain Yang and Ramanan's model, our model is still significantly better than their method. Observing the results obtained in Table 3, and by comparing our proposed method with the original DPM, both trained with the same range of images and tested with the same range of images, but a different one of trained images, we have improved the results with the proposed method by adding depth information, a KF, and using IKs to cut the number of points modeled in the DPM. Observing the results in Tables 2 and 3 and independently of the data set used to test or train parts, our proposed method obtains better solutions. This means that the results can be repeatable with different data sets.

In addition, in Table 3, our proposed method accuracy is compared both with and without a KF and obtained around 3.5% more accuracy using KF compared to not using KF. The reason for this is that when our proposed method fails in one frame, the wrong solutions obtained in the DPM are not corrected, while wrong solutions are corrected using the past information by KF when KF is employed.

Our results also show significant improvements over Kinect. However, this comparison is not completely fair since our method, having been trained on a smaller data set, is somewhat bias toward this data set. Thus, our results resemble a bias of our method toward the data set being trained on. Hence, if our method were to be tested on other data sets that have not been seen before, it would fail, whereas Kinect might not. This is possibly because Kinect has been trained on a much larger data set and its method can generalize better.

## Qualitative results

In this section, we analyze the qualitative results of our proposed method. Figure 7 shows the visual comparisons of our algorithm with the algorithms of Shotton et al.[35] (Kinect), Yang and Ramanan,[1] and Wang and Li.[2] The results of Wang do not seem better than those of Yang and Ramanan. The results of Yang and Ramanan and Kinect fail dismally when limbs fall outside the boundaries of the image or pose is more difficult. The Kinect algorithm also tends to fail when limbs fall outside boundaries and at times finds it difficult to identify the hip points that differ from person to person.

Our proposed method fails when two different joints are closer to each other, which could confuse our model with similar deformation and appearance costa for both joints (see Figure 7). Our proposed model could also fail when the pose configuration in question is not seen during training.

## Time complexity analysis

For our experiments, we use a system based on windows 7 with 64 bits and 4 GB RAM. The processor that we use is Inter Core Quad 2.33 GHz. For each frame, we calculate the average time taken by the proposed algorithm to process the frame. The used images have 320×240 pixels.

On training parts, our method takes about 8.12 min per frame, whereas Yang and Ramanan's method[1] takes about 8.54 min per frame, which is approximately a 5% gain in training time.

On testing part, our method takes about 7.26 s per frame using KF, whereas Yang and Ramanan's method[1] takes about 9.21 s per frame, which is approximately a 20% gain in pose estimation accuracy from Yang and Ramanan.[1]

Although the time performance of our method is much slower than Kinect, which is a real-time method, we show in our article that our method can be trained with fewer frames compared to Kinect, which requires hundreds of thousands of frames.

## Conclusions

In this article, we present a novel approach that combines monocular and depth information with a multichannel mixture of parts model, a novel structured LQE, and an IKs model to estimate joints for human pose estimation in RGBD data.

Our results demonstrate a significant improvement over state-of-the-art methods with CAD60 and our own data set. Our method can also be trained in less time and with a smaller fraction of training samples compared to the state-of-the-art.

### Declaration of conflicting interests

The author(s) declared no potential conflicts of interest with respect to the research, authorship, and/or publication of this article.

### Funding

The author(s) disclosed receipt of the following financial support for the research, authorship, and/or publication of this article: This work was partially financed by Plan Nacional de I+D, Comision Interministerial de Ciencia y Tecnologa (FEDERCICYT) under the project DPI2013-44227-R.